\documentclass{article}

\usepackage{arxiv}

\usepackage{fancyhdr}
\usepackage[utf8]{inputenc} % allow utf-8 input
\usepackage[T1]{fontenc}    % use 8-bit T1 fonts
\usepackage{hyperref}       % hyperlinks
\usepackage{url}            % simple URL typesetting
\usepackage{booktabs}       % professional-quality tables
\usepackage{amsfonts}       % blackboard math symbols
\usepackage{nicefrac}       % compact symbols for 1/2, etc.
\usepackage{microtype}      % microtypography
\usepackage{lipsum}
\usepackage{float}
\usepackage{graphicx}
\usepackage{multirow}
\graphicspath{ {./images/} }

\title{Abstractive Summarization of Low resourced Nepali language using Multilingual Transformers}

\author{
Prakash Dhakal\\
Department of Electronics and Computer Engineering\\ 
Institute of Engineering, Pulchowk Campus\\
Tribhuvan University, Nepal\\
\texttt{079msdsa016.prakash@pcampus.edu.np}\\
   \And
Daya Sagar Baral\\
Department of Electronics and Computer Engineering\\ 
Institute of Engineering, Pulchowk Campus\\
Tribhuvan University, Nepal\\
\texttt{dsbaral@pcampus.edu.np}\\
}

\begin{document}
\maketitle
\begin{abstract}
Automatic text summarization in Nepali language is an unexplored area in natural language processing (NLP). Although considerable research has been dedicated to extractive summarization, the area of abstractive summarization, especially for low-resource languages such as Nepali, remains largely unexplored. This study explores the use of multilingual transformer models, specifically mBART and mT5, for generating headlines for Nepali news articles through abstractive summarization. The research addresses key challenges associated with summarizing texts in Nepali by first creating a summarization dataset through web scraping from various Nepali news portals. These multilingual models were then fine-tuned using different strategies. The performance of the fine-tuned models were then assessed using ROUGE scores and human evaluation to ensure the generated summaries were coherent and conveyed the original meaning. During the human evaluation, the participants were asked to select the best summary among those generated by the models, based on criteria such as relevance, fluency, conciseness, informativeness, factual accuracy, and coverage. During the evaluation with ROUGE scores, the 4-bit quantized mBART with LoRA model was found to be effective in generating better Nepali news headlines in comparison to other models and also it was selected 34.05\% of the time during the human evaluation, outperforming all other fine-tuned models created for Nepali News headline generation.
\end{abstract}

% keywords can be removed

\section{Introduction}
The exponential growth of digital content, such as news articles, blogs, and social media, has made automatic text summarization a critical task in Natural Language Processing (NLP). This involves generating concise summaries that capture the main ideas of the original text while maintaining its meaning. Summarization is generally performed in two ways: extractive summarization and abstractive summarization. Abstractive summarization generates new sentences to convey the original text's meaning, requiring sophisticated language generation, while extractive summarization involves the extraction of key sentences or phrases from the original text. \\
Text summarization, particularly in generating news headlines, is vital for quickly conveying information and enabling further analysis like sentiment analysis and document classification. Efficient summarization is essential for managing the vast amount of digital text available today. While English text summarization has advanced significantly, there is a need for similar progress in low-resourced languages like Nepali.
The research work aimed to:
\begin{itemize}
    \item Assess the effectiveness of the mBART and mT5 model in performing abstractive summarization for Nepali language.
    \item Identify the key challenges and limitations in implementing mBART and mT5
models for abstractive summarization in low-resourced languages like Nepali.
    \item Determine the necessary modifications or enhancements to improve the performance of multilingual transformer models like mBART and mT5 for summarizing Nepali language texts.
    \item Explore how the developed abstractive summarization model can enhance the accessibility and usability of digital content in the Nepali language for various applications, such as news aggregation and educational resources.
\end{itemize}
In this research work, we have assessed the effectiveness of multilingual models: mBART\cite{mBART} and mT5\cite{mT5} in performing abstractive summarization specifically for Nepali News headline generation. For this, we have first created a summarization dataset through web scraping from various Nepali news portals. These multilingual models were then fine-tuned using different strategies. The performance of the fine-tuned models were then assessed using ROUGE scores and human evaluation to ensure the generated summaries were coherent and conveyed the original meaning.

\section{Related Work}
With the rise of transformer-based models \cite{vaswani2023attention}, various research works have been carried out using them for text summarization. Many studies focus on English, while research on the Nepali language is limited and primarily based on extractive summarization approaches.

\cite{RanabhatSenExt} introduced extractive summarization to produce summaries from multiple Nepali sentences by selecting a subset from the original text using TextRank \cite{mihalcea-tarau-2004-textrank}. These summaries contained the most important sentences of the input. They utilized TextRank for sentence scoring and topic modeling for summary evaluation.

\cite{MishraNewsSum} generated Nepali news headlines using GRU \cite{chung2014empirical} in an encoder-decoder fashion, taking the news content as input and generating a headline as output. The news was converted into word tokens and vectorized using FastText \cite{bojanowski2017enriching}, trained on a corpus of Nepali news articles and headlines collected from several web portals.

\cite{Khanalextra} employed an extractive method for Nepali text summarization using TextRanking \cite{mihalcea-tarau-2004-textrank} and LSTM \cite{LSTM}. They trained a Nepali news corpus with GloVe embeddings using different window sizes (10, 12, 15) and vector sizes (100, 200, 300). For extractive text summarization, they used Text Ranking and an attention-based LSTM model \cite{AttentionbasedLSTM}.

\cite{TimalsinaAttentionRNN} introduced an attention-based RNN for abstractive Nepali text summarization. They first created a Nepali text dataset by scraping Nepali news from online portals, then designed a deep learning-based summarization model using an encoder-decoder recurrent neural network with attention. Specifically, Long Short-Term Memory (LSTM) \cite{LSTM} cells were used in both the encoder and decoder layers. They built nine models by varying hyperparameters and reported Recall-Oriented Understudy for Gisting Evaluation (ROUGE) scores \cite{ROUGE} to evaluate performance.

\section{Methodology}
\subsection{Data Collection}
\begin{figure}[h] % Use [p] for placing the figure on a separate page
    \includegraphics[width=1\textwidth]{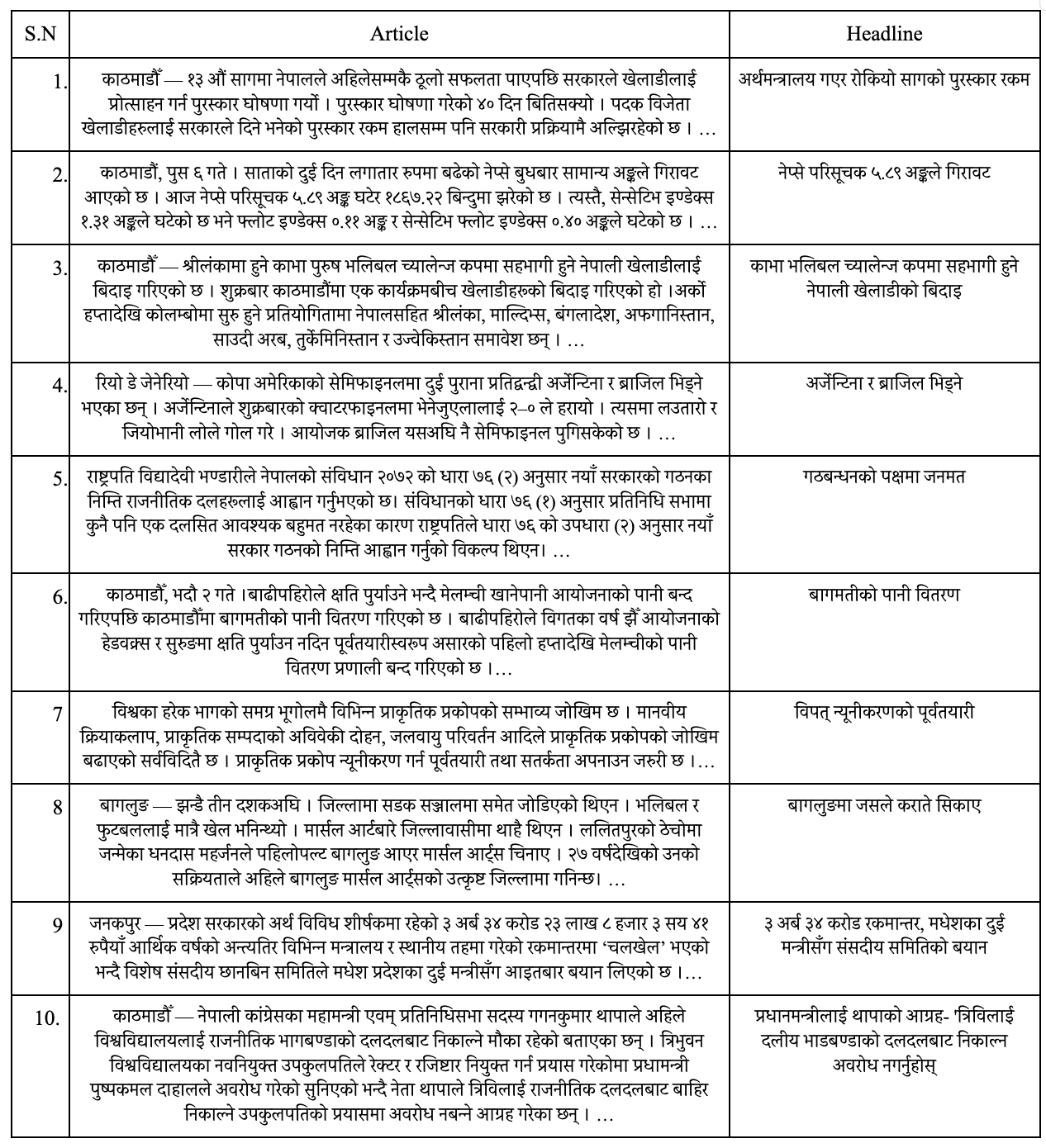} % Replace
    \centering
    \caption{Data Sample}
    \label{fig:data sample}
\end{figure}

A comprehensive dataset of Nepali news articles, was created with web scraping from various online news portals like BBC Nepali\cite{hasan-etal-2021-xl}, Kantipur and Gorkhapatra. For web scraping, libraries like \textit{BeautifulSoup} and \textit{Selenium} were used. Data in each news portals were in different format and different strategies had to be adopted to extract data from them. Running a single script to collect the data would have taken forever to get the complete dataset, so in order to expedite the data collection process, various parallel processing techniques were adapted. A sample of dataset obtained from this process has been presented in Figure~\ref{fig:data sample}.
\subsection{Data Preprocessing}
In this step, we have removed HTML tags, special characters, and irrelevant sections of the text (such as advertisements and navigation links). As the data was collected in two steps, the headlines and their corresponding article bodies had to be joined to create the complete dataset.\\
The collected dataset still had numerous characters that were not part of the Nepali Devanagari Character Set. These extraneous characters would have degraded the overall text quality and negatively impacted model performance. Specifically, the unwanted characters include Latin letters (a-z, A-Z), Arabic numerals (0-9), etc. To mitigate these issues, these characters have been removed from the dataset.
A prefix was added to the input text to indicate the summarization task to the model and it helped the model to better understand the context and the task it needs to perform.\\
The input texts (articles) and the target texts (headlines) were then, tokenized to a maximum length of 1024 and 20 tokens respectively, ensuring that longer texts were truncated.
The tokenized headlines from the previous step were then, set as labels in the model inputs. This helped the model to learn the mapping from the input text to the target headlines during training.
A data collator was used after the tokenization, in order to dynamically pad the inputs and labels to the maximum length during the batching process, ensuring the efficient utilization of the model's capabilities.

\subsection{Exploratory Data Analysis}
The dataset, meticulously compiled from various news portals, encapsulated a total of 70,769 articles, categorized into ten distinct thematic areas: News, Sports, Opinion, Entertainment, Feature, Diaspora, World, Education, Blog and Others(Mix). The dataset had more data related to News category, while blog category had the least amount of data. The dataset were, then splitted into training, validation, and test sets in an 70-20-10 ratio to ensure robust model evaluation.
\begin{table}[H]
    \centering
    \caption{Data distribution in training and testing dataset}
    \label{table:data_distribution}
    \begin{tabular}{|c|c|c|}
        \hline
        \textbf{Dataset type} & \textbf{Count} \\
        \hline
        Training Set &  49,538\\
        \hline
        Validation Set & 14,154\\
        \hline
        Test Set & 7,076 \\
        \hline
    \end{tabular}
    \label{tab:sample_table}
\end{table}

\begin{table}[H]
    \centering
    \caption{Data Statistics}
    \label{table:Data Statistics}
    \begin{tabular}{|c|c|c|}
        \hline
        \textbf{S.N} & \textbf{Category} & \textbf{Count} \\
        \hline
        1 & News & 36798 \\
        \hline
        2 & Sports & 18767 \\
        \hline
        3 & Others(Mix) & 7258 \\
        \hline
        4 & Opinion & 2358 \\
        \hline
        5 & Entertainment & 2144 \\
        \hline
        6 & Feature & 2014 \\
        \hline
        7 & Diaspora & 750 \\
        \hline
        8 & World & 462 \\
        \hline
        9 & Education & 188 \\
        \hline
        10 & Blog & 30 \\
        \hline      
         & Total & 70,769 \\
        \hline   
    \end{tabular}
    \label{tab:sample_table}
\end{table}

\subsection{Model Selection and Fine-Tuning}
\subsubsection{Model Selection}
For summarization task in other languages, the  multilingual transformer-based models, specifically \textit{mBART}\cite{tang2020multilingual} and \textit{mT5}\cite{mT5} have shown impressive results. These models have already been pretrained on vast amounts of multilingual text, which gives them a strong foundation for understanding complex linguistic structures in Nepali. This helps in generating accurate and coherent summaries without needing massive language-specific datasets. Both models offer flexibility in fine-tuning, allowing us to adapt them specifically to the nuances of Nepali text summarization. This ability helps improve performance in low-resource settings where language-specific models are not readily available. There were different variants of mBART and mT5 available for the summarization task, but the mBART-large-50 and mT5-base models in particular had only 600M and 598M trainable parameters respectively making it suitable for our use case. If we had decided to go with other variants of mBART and mT5, it would have been computationally expensive and time consuming and our existing free resources would have been incapable of doing this task. Due to this reason, we had to go with these two variants which were computationally less expensive in comparison to the other variants. Training all of the parameters of these models were still computationally expensive and time consuming, so in order to reduce the number of trainable parameters in the model, LoRA\cite{hu2021lora} was used along with quantization technique as suggested in QLoRA\cite{QLora}.
\\
\subsubsection{Fine-Tuning}
To enhance efficiency, we stored the dataset on Hugging Face. During the fine-tuning process, the model weights and configurations obtained after each training session were pushed to Hugging Face for every model. Given the substantial computational demands of fine-tuning our language models, we found Kaggle to be the most suitable platform. It offered free access to the NVIDIA TESLA P100 GPU, allowing us to conduct uninterrupted training sessions for up to 12 hours.
 
The following training arguments were set in the trainer and in the LoRA for the training in each models:\\
\begin{table}[H]
    \centering
    \caption{Training arguments for trainer}
    \label{table:training_setup_for_trainer}
    \begin{tabular}{|c|c|c|}
        \hline
        \textbf{Parameters} & \textbf{Value}\\
        \hline
        evaluation\_strategy &  epoch\\
        \hline
        learning\_rate & 5e-4\\
        \hline
        per\_device\_train\_batch\_size & 5 \\
        \hline
        per\_device\_eval\_batch\_size & 5 \\
        \hline
        weight\_decay & 0.01\\
        \hline
        num\_train\_epochs & 3\\
        \hline
        per\_device\_train\_batch\_size & 5 \\
        \hline
    \end{tabular}
\end{table}
\begin{table}[H]
    \centering
    \caption{Training arguments for LoRA}
    \label{table:training_arguments_for_LoRA}
    \begin{tabular}{|c|c|c|}
        \hline
        \textbf{Parameters} & \textbf{Value}\\
        \hline
        r &  32\\
        \hline
        lora-alpha & 32\\
        \hline
        lora-dropout & 0.1 \\
        \hline
        bias & lora\_only \\
        \hline
    \end{tabular}
\end{table}

The pre-trained models were then, adapted using the LoRA configuration. This involved updating the model's weights based on the low-rank adaptations, making it more efficient for the specific task of Nepali news headline generation. Finally, the adapted model were fine-tuned using the same training process as described earlier. The low-rank update enabled faster and more efficient training, resulting in a model that could generate high-quality headlines.

\subsection{Evaluation}
The evaluation strategy was set to run at the end of each epoch, allowing for periodic assessment of the model's performance during training. A custom function to compute evaluation metrics was provided to the trainer. This function calculated ROUGE scores to evaluate the quality of the generated headlines. The model's performance was finally assessed on the testing set using the custom evaluation function and helped in understanding the model's ability to generate accurate and coherent headlines from Nepali news articles.\\
To assess the models' performance, a survey was conducted with 62 participants fluent in Nepali. They were asked to evaluate summaries of 10 different sentences from various categories, each one having a summary generated from six different models ensuring fairness in the evaluation and were also asked to do it based on relevance, fluency, conciseness, informativeness, factual accuracy, and coverage of the summary.

\section{Experimental Setup}
For the execution of this experiment, the following setup was created:
\subsection{Environment Configuration:}
\subsubsection{Hardware Setup:} 
Given the substantial computational demands of fine-tuning
our language models, we found Kaggle to be the most suitable platform. It offered free access
to the NVIDIA TESLA P100 GPU (16GB), allowing us to conduct uninterrupted training sessions
for up to 12 hours. For storing the data, the model weights and the configurations obtained after each training session, Hugging Face was used. \\
\subsubsection{Software Environment:} 
The experiments were ran using Python 3.12.3 along with key libraries such as PyTorch, BeautifulSoup, Selenium, Pandas, Numpy, Matplotlib, Plotly etc.\\
\subsection{Experimental Workflow:}
\subsubsection{Dataset Handling:} 
The dataset was processed in batches during training, with each batch containing 10k approx. samples. A total of 50k and 14k news articles and their corresponding summaries were used in this experiment for training and validation respectively. The dataset was fully loaded into the memory for each model during the start of training.\\
\subsubsection{Batch Processing:} 
Batch processing was implemented to streamline training and evaluation. Training was performed with a batch size of 5 and ran for 3 epochs and validation was carried out at regular intervals to track performance improvements.\\
\subsubsection{Training Time:} 
The total training time per model was approximately 12 hours.\\
\subsubsection{Hyperparameter Settings:}
The key hyperparameters used were: learning rate = 5e-4, weight decay = 0.01. These parameters were optimized to balance model convergence and training stability.\\
\subsection{Evaluation Setup:}
\subsubsection{Automated Evaluation:} 
Evaluation metrics, such as ROUGE, were computed using the Rouge library. The results were automatically logged and stored for further analysis. The evaluation was carried out on the validation dataset after each epoch and final evaluation was done on the test dataset at the end.\\
\subsubsection{Human Evaluation:} 
During human evaluation, human evaluators were asked to select the best summary among different summaries generated from different models for different sentences based on factors such as relevance, fluency, and informativeness. A simple Google form was created and used to streamline the collection of feedback, ensuring that responses were gathered efficiently.\\
\section{Results}
The overall ROUGE scores in terms of precision, recall, and F1-scores for all the models are presented in Table \ref{table:ROUGE_score_of_different_models}. These scores provide a comprehensive summary of the experiments. Based on the results in the Table \ref{table:ROUGE_score_of_different_models}, 4-bit quantized mBART with LoRA has outperformed all other models in terms of ROUGE scores.\\ 
The results obtained from the human evaluation for model comparison are also presented in Table \ref{table:human_evaluation}. From this table, we observed that the summarization generated from 4-bit quantized mBART with LoRA were selected maximum number of times supporting the results from the automatic evaluation and also indicating that the output generated from this model were more relevant, fluent, concise, informative, factually accurate and had better coverage in comparison to other models. \\
However, models like the 4-bit and 8-bit quantized mT5 failed to generate sentences properly. There are several factors that could explain this poor performance. Firstly, the quantization process might have adversely affected the mT5 model more than the mBART model, leading to a loss in precision and overall capability to generate coherent summaries. Secondly, the mT5 model may require more extensive fine-tuning to adapt to the specifics of the Nepali language, which was not sufficiently covered in this experiment. Additionally, the architecture of mT5 might inherently be less robust to quantization techniques compared to mBART, which could explain the significant drop in performance. These factors combined resulted in the mT5 models underperforming in this task.
\\

\begin{table}[H]
\footnotesize % Reduced font size
\setlength{\tabcolsep}{10pt} % Adjust space between columns
    \centering
    \resizebox{1\textwidth}{1\height}{ % Adjust width and height scaling here
      \begin{tabular}{|l|c|c|c|c|c|c|c|c|c|}
        \hline
        {\textbf{Model}} &
          \multicolumn{3}{c|}{\textbf{ROUGE-1}} &
          \multicolumn{3}{c|}{\textbf{ROUGE-2}} &
          \multicolumn{3}{c|}{\textbf{ROUGE-L}} \\
        \cline{2-10}
          & \textbf{P} & \textbf{R} & \textbf{F1} & \textbf{P} & \textbf{R} & \textbf{F1} & \textbf{P} & \textbf{R} & \textbf{F1} \\
        \hline
        mBART+LoRA & 0.3797 & 0.3517 & 0.355 & 0.211 & 0.196 & 0.1964 & 0.3684 & 0.3411 & 0.3443 \\
        \hline
        \textbf{4-bit quantized mBART+LoRA} & 0.3865 & \textbf{0.354} & \textbf{0.359} & \textbf{0.2163} & \textbf{0.1984} & \textbf{0.1999} & \textbf{0.3754} & \textbf{0.344} & \textbf{0.3488} \\
        \hline
        8-bit quantized mBART+LoRA & \textbf{0.3871} & 0.35 & 0.3574 & 0.2141 & 0.1941 & 0.1969 & 0.3754 & 0.3395 & 0.3466 \\
        \hline
        mT5+LoRA & 0.335 & 0.3248 & 0.3218 & 0.1746 & 0.1701 & 0.1675 & 0.3252 & 0.3154 & 0.3123 \\
        \hline
      \end{tabular}
    }
    \caption{ROUGE scores of different models on the test dataset}
    \label{table:ROUGE_score_of_different_models}  % Label goes after the caption
\end{table}

\begin{figure}[h]
    \centering
    \includegraphics[width=0.65\textwidth, height=0.3\textheight]{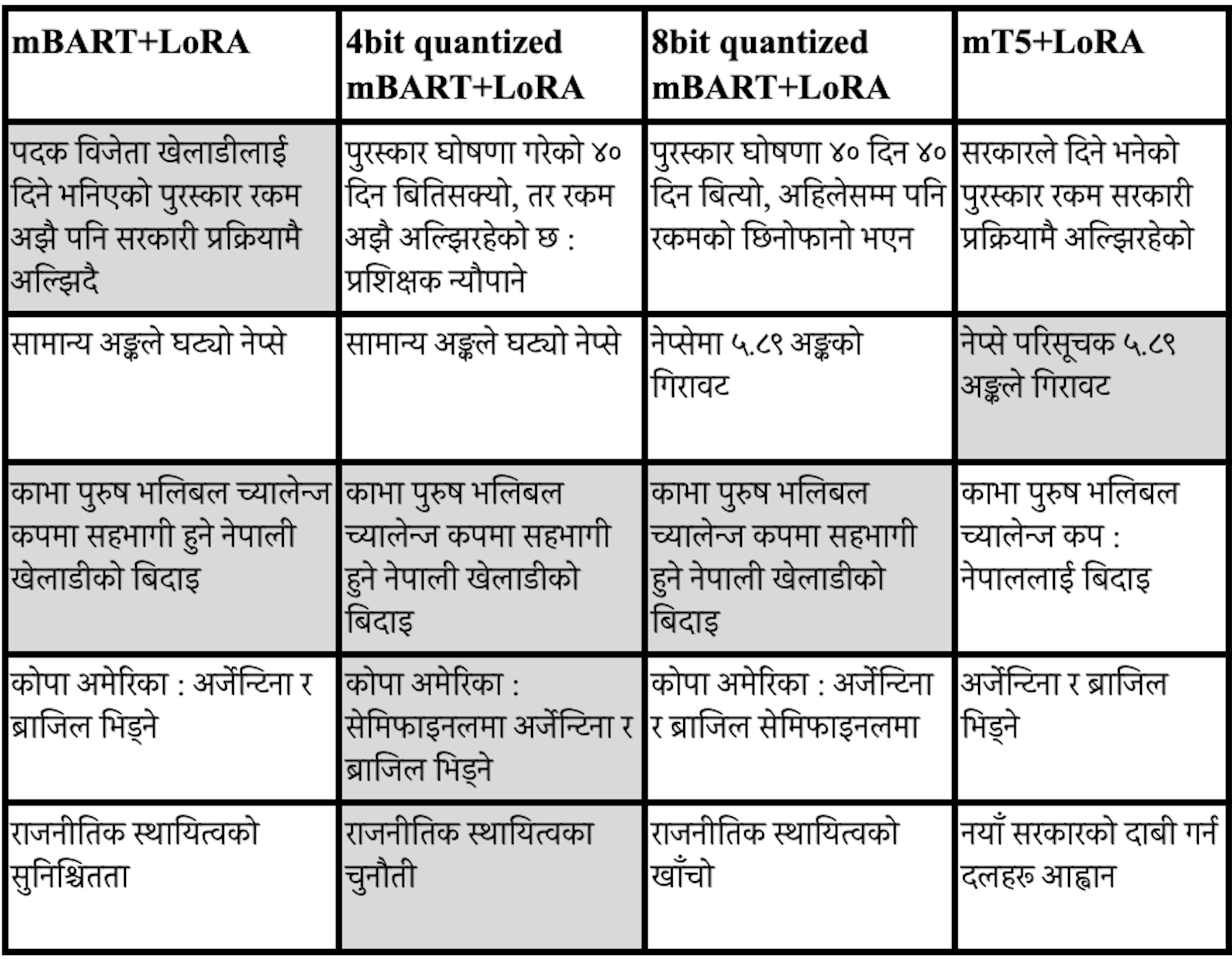} 
    \caption{Summaries generated by different models (1-5)} 
    \textit{Note: The highlighted entries in the above table received the maximum number of votes in the survey.}
    \label{fig:out15}
\end{figure}

\begin{figure}[H]
    \centering
    \includegraphics[width=0.65\textwidth, height=0.3\textheight]{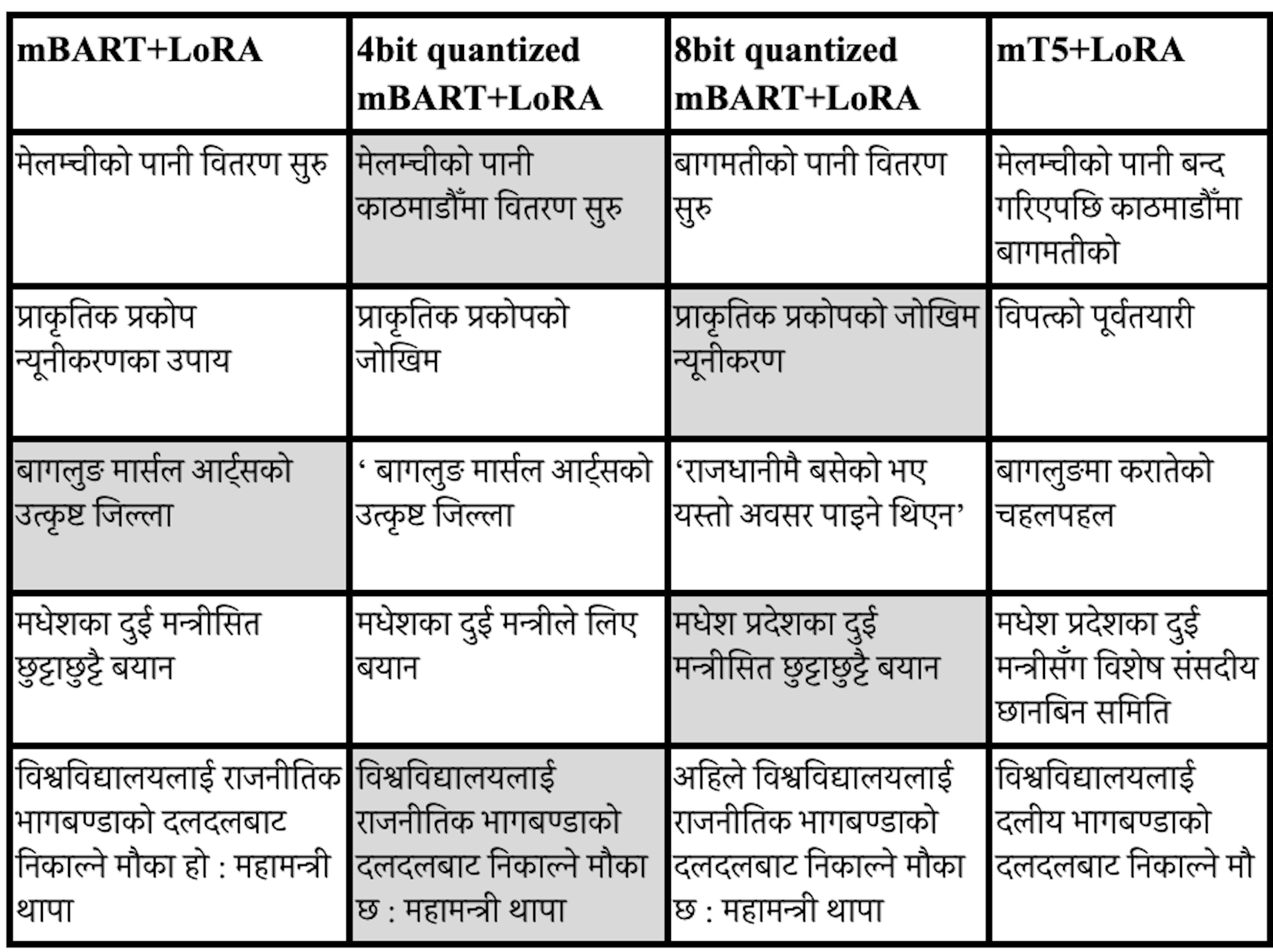} 
    \caption{Summaries generated by different models (6-10)} 
    \textit{Note: The highlighted entries in the above table received the maximum number of votes in the survey.}
    \label{fig:out610}
\end{figure}

\begin{table}[H]
    \centering
    \caption{Results from the Human Evaluation}
    \label{table:human_evaluation}
    % \resizebox{0.6\textwidth}{0.8\height}{ % Adjust width and height scaling here
    \begin{tabular}{|c|c|c|c|}
        \hline
        \textbf{Model} & \textbf{Number of votes received} & \textbf{Percentage of votes (\%)} \\
        \hline
        4bit quantized mBART+LoRA &  235 & 34.06 \\
        \hline
        8bit quantized mBART+LoRA & 191 & 27.68 \\
        \hline
        mBART+LoRA & 164 & 23.77 \\
        \hline
        mT5+LoRA & 100 & 14.49 \\
        \hline
        4bit quantized mT5+LoRA & 000 & 00.00 \\
        \hline
        8bit quantized mT5+LoRA & 000 & 00.00 \\
        \hline
    \end{tabular}
    % End resizebox
\end{table}

\begin{figure}[H]
    \centering
    \includegraphics[width=0.75\textwidth, height=0.3\textheight]{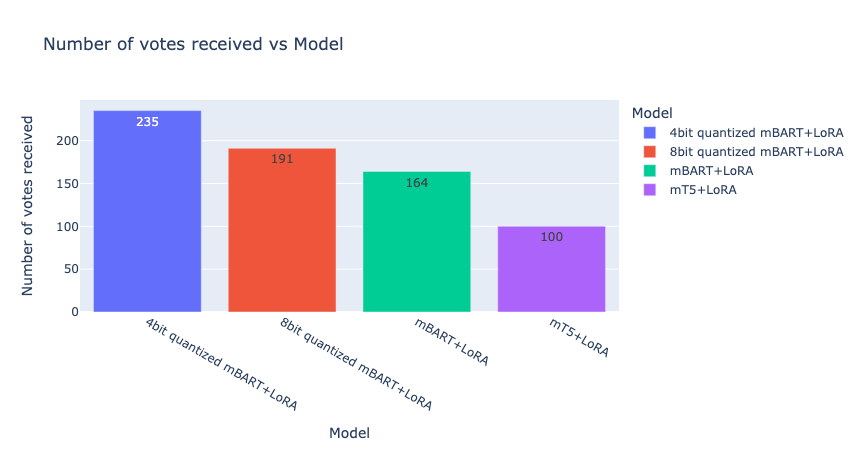} % Adjust both width and height as needed
    \caption{Results from the Human Evaluation}
    \label{fig:out61}
\end{figure}
% \clearpage
\section{Conclusion}
This research focused on enhancing the generation of Nepali news headlines using advanced summarization models. In this study, a diverse dataset of Nepali news articles was successfully collected and preprocessed to ensure its readiness for model training. By leveraging state-of-the-art multilingual models, such as mBART and mT5, and incorporating techniques like LoRA and quantization, these models were trained and compared with automatic evaluation technique like ROUGE and finally with human evaluation to conclude the comparison.\\
The evaluation results revealed that the 4-bit quantized mBART model with LoRA achieved the highest performance in generating accurate and coherent headlines. This model outperformed the other variations, including the quantized mT5 models, which struggled with performance issues likely due to the effects of quantization and the model's adaptation to the Nepali language.\\
Overall, this study demonstrates the effectiveness of mBART with LoRA in the context of Nepali news headline generation and highlights the potential areas for improvement in models like mT5. The findings underscore the importance of optimizing models and techniques to achieve high-quality summarization in diverse linguistic contexts.\\
To further enhance the effectiveness of summarization models, several recommendations can be considered. First, improving the performance of the mT5 model is essential. This involves exploring ways to handle quantization more effectively, possibly by experimenting with different strategies or refining the model’s fine-tuning process to better adapt to the Nepali language.\\
Extending the training and fine-tuning phases could be one of the ways to improve the performance of model. This approach would involve using additional data and applying diverse summarization tasks to strengthen the model's adaptability and overall performance. Additionally, exploring alternative multilingual transformer models or newer architectures could offer valuable insights and alternative methods for summarization, potentially revealing more effective solutions than those currently utilized.\\ 
Refining quantization techniques is another crucial area of focus. A deeper investigation into how different quantization methods impact model performance can lead to improved efficiency without compromising quality. This may include experimenting with varying quantization levels or employing mixed-precision approaches.

% \FloatBarrier % Ensure all floats before this point are processed
\section*{Acknowledgment}
I would like to express my deepest gratitude to the Department of Electronics and Computer Engineering, Institute of Engineering, Pulchowk Campus, Tribhuvan University, Nepal for providing me oppportuity to undertake this research work. 
A special thanks to my supervisor, Asst. Prof. Daya Sagar Baral, whose expert guidance, unwavering support, and constructive feedback have been instrumental in the successful completion of this work. I am also grateful to Asst. Prof. Dr. Aman Shakya, the program coordinator for the M.Sc. in Computer Engineering (Specialization in Data Science and Analytics), for his invaluable assistance and insightful advice throughout the research. Furthermore, I extend my sincere thanks to Assoc. Prof. Dr. Arun Timalsina, Asst. Prof. Dr. Basanta Joshi, and Asst. Prof. Sanjivan Satyal from Pulchowk Campus for their valuable feedback and guidance during the course of this study and in the preparation of the report. Lastly, I would like to thank my friends and family for their continuous support and encouragement.

\bibliographystyle{unsrt}  
\bibliography{main}  %%% Remove comment to use the external \bibliographystyle{IEEEtran}

\section*{Appendices}
\subsection{Dataset Details}
\begin{itemize}
    \item Data Source: Data scrapped from the different web portals, which were used for the training of the models is available in the following link. \\
    (Link: https://www.kaggle.com/datasets/dhakal2444/nepali-news-dataset/data)
\end{itemize}
\subsection{Additional Results}
\begin{itemize}
    \item Model Performance: More detailed ROUGE score of each model obtained during training are presented below:

    \begin{figure}[h]
    \centering
    \includegraphics[width=1\textwidth, height=0.1\textheight]{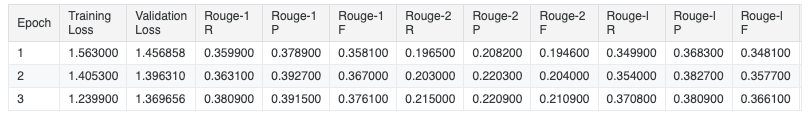} % Adjust both width and height as needed
    \caption{ROUGE scores of mBART with LoRA obtained during training}
    \label{fig:eval_mBART}
    \end{figure}

    \begin{figure}[h]
    \centering
    \includegraphics[width=1\textwidth, height=0.1\textheight]{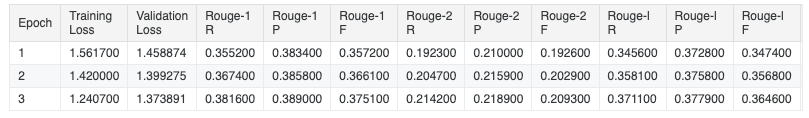} % Adjust both width and height as needed
    \caption{ROUGE scores of 4bit mBART with LoRA obtained during training}
    \label{fig:eval_4bitmBART}
    \end{figure}
    
    \begin{figure}[h]
    \centering
    \includegraphics[width=1\textwidth, height=0.1\textheight]{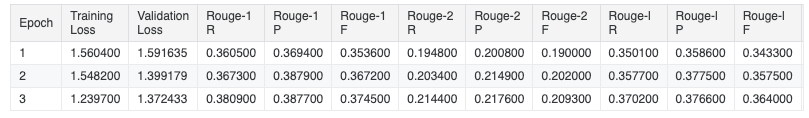} % Adjust both width and height as needed
    \caption{ROUGE scores of 8bit mBART with LoRA obtained during training}
    \label{fig:eval_8bitmBART}
    \end{figure}
        
    \begin{figure}[H]
    \centering
    \includegraphics[width=1\textwidth, height=0.1\textheight]{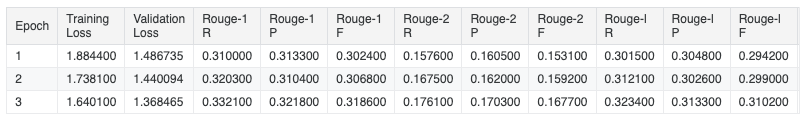} % Adjust both width and height as needed
    \caption{ROUGE scores of mT5 with LoRA obtained during training}
    \label{fig:eval_mT5}
    \end{figure}

    \item Human Evaluation: The snippets of the Google form created for the human evaluation is presented below:
    
    \begin{figure}[h]
    \centering
    \includegraphics[width=0.55\textwidth, height=0.45\textheight]{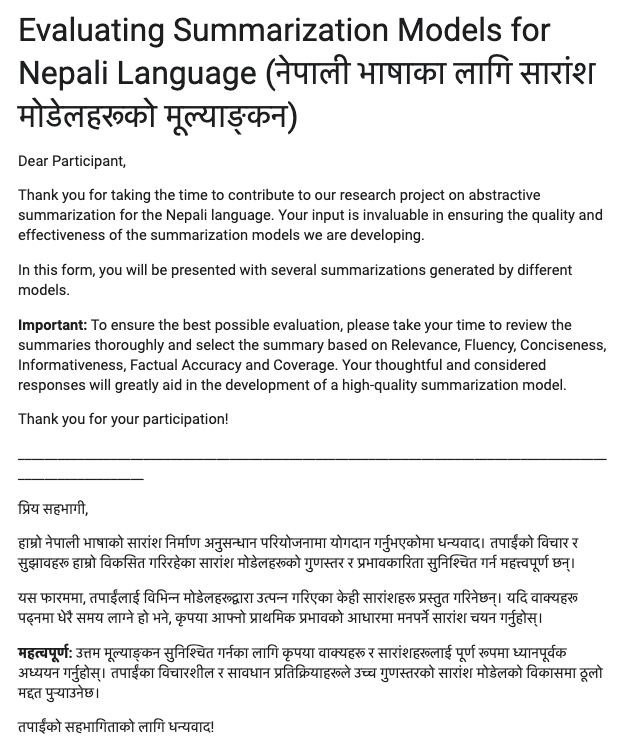} % Adjust both width and height as needed
    \caption{Snippet of the Google Form used for human evaluation}
    \label{fig:eval_hu7man}
    \end{figure}
        
    \begin{figure}[H]
    \centering
    \includegraphics[width=0.55\textwidth, height=0.45\textheight]{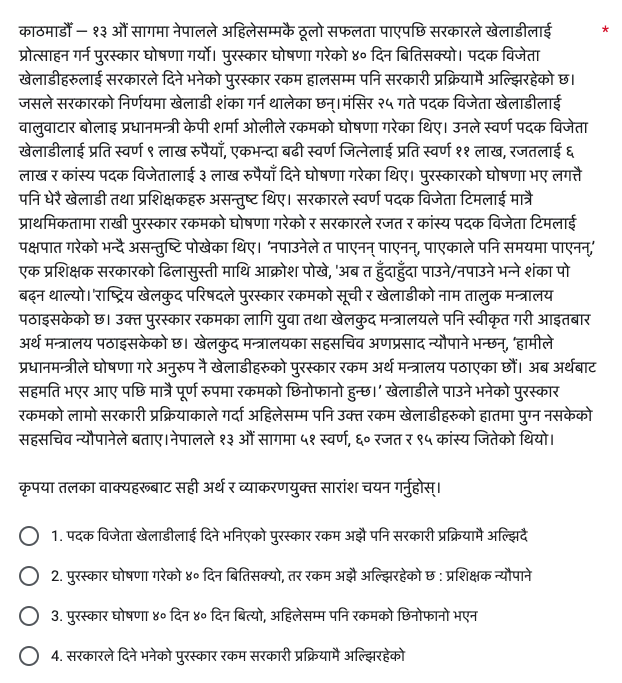} % Adjust both width and height as needed
    \caption{Snippet of the sentence and summaries used during human evaluation}
    \label{fig:eval_humna}
    \end{figure}

\end{itemize}

\end{document}